\documentclass[11pt, a4paper]{article}
\usepackage[a4paper, top=2.5cm, bottom=2.5cm, left=2cm, right=2cm]{geometry}
\usepackage[english]{babel}

\usepackage{booktabs}
\usepackage{tabularx}
\usepackage{graphicx}
\usepackage{tikz}
\usetikzlibrary{positioning, arrows.meta, shapes.geometric, calc}
\usepackage{hyperref}
\usepackage{enumitem}

\title{\textbf{Beyond Vector Similarity: Hierarchical Context-Aware Graph RAG vs Standard RAG in Enterprise Code Migration}}

\author{
  Nilesh Jaiswal, Arjit Shukla, Divya Malhotra, Aniket Agrawal, \\
  Saurabh Garg, Suddhasatwa Bhaumik, Suchit Puri \\
  \texttt{\{nileshjaiswal, arjitshukla, malhotradi, aniketagrawal,} \\
  \texttt{saurabhkgarg, suddhasatwa, suchitpuri\}@google.com} \\
  Google Cloud
}

\date{April 6, 2026}

\begin{document}

\maketitle

\begin{abstract}
As enterprises modernize legacy systems (e.g., monolithic Java architectures to Python microservices), Large Language Models (LLMs) have become instrumental in automated code translation. 

However, traditional vector-based Retrieval-Augmented Generation (Standard RAG) struggles with topological relationships, fetching isolated text chunks that frequently sever inheritance chains and lead to high compilation failure rates. 

This paper presents a comparative analysis between Standard RAG and a novel \textbf{Hierarchical Context-Resident Graph (HCRG)} methodology. Our pipeline utilizes tree-sitter for polyglot Abstract Syntax Tree (AST) extraction, mapping architectural edges into a Google Cloud Spanner Property Graph, and serializing this structure into a Gemini (on Vertex AI) Context Cache to enable topological, parent-first code translation.

By shifting evaluation from naive text-overlap to a custom 7-metric framework measuring Software Engineering (SE) utility, empirical evaluations on the \texttt{spring-petclinic-genai} repository demonstrate significant structural improvements. \textbf{Graph RAG decisively mitigates dependency loss, dropping the API hallucination rate from 56.4\% to 16.2\%}. 

Furthermore, it \textbf{improves Dependency Resolution Quality (DRQ) from 34.8\% to 65.9\%} and \textbf{enhances Parent-Child Consistency (PCC) from 26.7\% to 45.5\%}. Interestingly, traditional lexical metrics fail to capture this divergence; \textbf{both methodologies achieved an identical 91\% average CodeBLEU score}, effectively masking Standard RAG’s structural failures behind syntactically plausible but broken code.

However, the results indicate that \textbf{Graph RAG is not strictly superior across all dimensions}. Providing the LLM with dense, global structural context introduces new vulnerabilities: \textbf{Graph RAG suffers a severe degradation in Cyclomatic Complexity Consistency (dropping from Standard RAG's 71.6\% to 46.7\%) due to defensive over-engineering by the LLM}, alongside a \textbf{slight drop in Docstring Preservation (67.0\% down to 61.0\%) caused by prompt attention dilution}. 

Ultimately, this research validates that while Graph RAG trades an increase in code complexity for critical reductions in API hallucinations, it offers a substantially more viable and architecturally sound path for automated enterprise codebase modernization.
\end{abstract}

\section{Introduction}
The migration of monolithic, legacy codebases to modern microservices requires far more than line-by-line syntax translation; it demands an understanding of deeply nested architectural intent. Enterprise software is highly coupled, with single business operations spanning across multiple files, interfaces, and base classes. While Large Language Models (LLMs) excel at localized code generation, standard "Chat with your Codebase" paradigms utilizing Vector Retrieval-Augmented Generation (Standard RAG) are inherently limited by the semantic boundaries caused by the text chunking process. When an LLM attempts to translate a child class, semantic retrieval often fails to fetch the inherited parent class or distant utility interfaces unless they share high lexical overlap. This structural blindness severs topological relationships, leading to high compilation failure rates, missing imports, and severe API hallucinations.

Graph-based Retrieval-Augmented Generation (Graph RAG) addresses this limitation by modeling software as a deterministic network rather than a collection of floating text embeddings. Recent literature highlights that AST-derived knowledge graphs provide superior explainability and traceability over pure semantic retrieval \cite{chinthareddy2026, dong2026, edge2024}. To advance this paradigm, this paper presents the engineering and evaluation of a Context-Aware Migration Pipeline powered by a novel Hierarchical Context-Resident Graph (HCRG) methodology. Utilizing \texttt{tree-sitter}, we deterministically extract polyglot Abstract Syntax Trees (AST) and map their architectural dependencies (e.g., \texttt{INHERITS}, \texttt{CALLS}, \texttt{IMPORTS}) into a Google Cloud Spanner Property Graph. Rather than treating this structure merely as data to be queried, the HCRG pipeline serializes the repository’s "Architecture Skeleton" and caches it within the LLM's working memory via Vertex AI Context Caching. This globally aware state allows the pipeline to orchestrate a topologically sorted migration—translating parent classes before their children—ensuring that the LLM is contextually grounded by the newly generated Python parent code rather than relying on isolated semantic snippets.

To rigorously evaluate the efficacy of this AST-derived Graph RAG pipeline against a Standard RAG baseline, we perform an automated Java-to-Python migration on the highly coupled \texttt{spring-petclinic-genai} repository. Crucially, we demonstrate why traditional LLM evaluation metrics fail to capture the architectural superiority of Graph RAG. Relying primarily on lexical N-gram overlap, both Standard RAG and Graph RAG achieved an identical 91\% average CodeBLEU score in our experiments, creating a false equivalence that masked underlying structural failures. 

To uncover the true viability of the generated code, we introduce a detailed, AST-based evaluation framework comprising seven Software Engineering (SE) metrics: Code Migration Quality (CMQ), Parent-Child Consistency (PCC), Type Hint Completion (THC), Linter Status (StaQ), Dependency Resolution Quality (DRQ), Cyclomatic Complexity Consistency (CCC), and Docstring Preservation (DP). 

Through this rigorous programmatic evaluation, we reveal that while Standard RAG generates syntactically plausible code, it hallucinates APIs and dependencies in over half of its outputs (56.4\% hallucination rate). Conversely, Graph RAG leverages global cross-file resolution lookups to drop the true hallucination rate to 16.2\%, while dramatically improving Dependency Resolution Quality (from 34.8\% to 65.9\%) and Parent-Child Consistency (from 26.7\% to 45.5\%). However, this research does not conclude that Graph RAG is uniformly superior. By providing the LLM with a comprehensive "god's-eye view" of the repository structure, Graph RAG triggers defensive over-engineering, resulting in bloated Cyclomatic Complexity and slight degradations in prompt attention toward secondary tasks like Docstring Preservation. This paper details the architecture of the HCRG pipeline, dissects these empirical trade-offs, and provides a blueprint for leveraging deterministic graphs in enterprise code modernization.

It is important to note that while this research leverages Google Cloud Spanner and Gemini 2.5 Pro Vertex AI for the experimental reference implementation, the underlying HCRG methodology is fundamentally platform-agnostic. The core architectural principles—deterministic AST ingestion, graph-based topological sorting, and context-resident caching—are equally applicable and deployable across any standard property graph database and comparable generative AI services, whether hosted on-premise or within alternative cloud environments.

\section{Glossary of Terms}

To ensure clarity throughout this comparative analysis, we define the core underlying mechanisms, methodologies, and metrics referenced in our evaluation pipeline:

\begin{description}[style=nextline, leftmargin=1cm]

\item[Abstract Syntax Tree (AST)] 
A highly structured, hierarchical tree representing the syntactic structure of source code. Unlike naive regex matching, an AST allows us to deterministically extract structural boundaries (Classes, Function Definitions, Arguments) and accurately capture the semantic intent of the original code. In our ingestion pipeline, we utilize \texttt{tree-sitter} to perform robust, polyglot AST parsing across Java, Python, and Go.

\item[Standard RAG (Vector Search)] 
A retrieval framework that splits text into arbitrary chunks (e.g., using recursive character splitters), embedding them into a mathematical high-dimensional vector space. Retrieval relies entirely on Cosine Similarity against the user's prompt. It is frequently inadequate for code migration because a deeply nested child class might share zero lexical or semantic similarity with the crucial interface it inherits, thereby severing topological relationships.

\item[Graph RAG] 
An advanced RAG framework that maps text and topological relationships into a deterministic graph database. This allows for precise, multi-hop retrieval of context based on strict entity relationships (e.g., explicitly traversing a graph edge to fetch a target class, the specific interface it implements, and the remote methods it calls), drastically reducing the hallucination of non-existent APIs.

\item[Hierarchical Context-Resident Graph (HCRG)] 
The novel architectural methodology introduced in this paper. Instead of treating the LLM solely as a reasoning engine and the graph solely as isolated storage, HCRG serializes the repository's highest-level structure into the LLM's active working memory via Vertex AI Context Caching. This enables zero-latency traversals for architectural reasoning, while dynamically falling back to the graph database for deep AST chains.

\item[Architecture Skeleton]
A condensed, serialized JSON representation of the entire repository's "Hot Architecture" (including classes, parent hierarchies, and method signatures). This skeleton is constructed via graph queries and pushed into the Context Cache to provide the LLM with a low-latency, "god's-eye view" of the codebase prior to generation.

\item[Directed Acyclic Graph (DAG)] 
A mathematical topological structure with no directed cycles. In our pipeline, we convert the codebase's inheritance trees into a DAG to dictate the exact chronological order in which files must be migrated (e.g., ensuring \texttt{BaseEntity} is translated before \texttt{Person}, and \texttt{Person} before \texttt{Vet}). This ensures that newly generated Python parent code acts as strict contextual grounding when generating child classes.

\item[Spanner Property Graph] 
Google Cloud Spanner's globally distributed database engine, utilized in this architecture to natively unify relational tabular data with graph data. It stores logical AST nodes (e.g., \texttt{CodeFiles}, \texttt{CodeStructures}, \texttt{Callables}), native \texttt{ARRAY<FLOAT64>} embeddings for vector search, and edges (e.g., \texttt{INHERITS}, \texttt{CALLS}, \texttt{IMPORTS}), all queried via ISO-standard Graph Query Language (GQL).

\item[Software Engineering (SE) Utility Metrics]
Our custom programmatic evaluation framework designed to supersede generic lexical metrics like CodeBLEU. It comprises seven deterministic metrics tailored for migration viability: Code Migration Quality (CMQ), Parent-Child Consistency (PCC), Type Hint Completion (THC), Linter Status (StaQ), Dependency Resolution Quality (DRQ), Cyclomatic Complexity Consistency (CCC), and Docstring Preservation (DP).

\end{description}

\section{Architectural Paradigms \& Diagrams}

\subsection{Baseline Methodology: Standard RAG (Vector Search)}

To establish a comparative baseline, we implemented a traditional Retrieval-Augmented Generation (Standard RAG) pipeline that treats the codebase as a flat text corpus, prioritizing semantic similarity over topological relationships. Target source files (.java and .proto) are processed using a naive recursive character splitter. These isolated chunks are embedded and indexed in a Vector Database. The fundamental limitation of this approach is assuming semantic similarity equates to structural relevance; because it relies on lexical overlap, the pipeline often retrieves parallel sibling classes rather than the actual parent classes or distant interfaces needed for translation, acting as the primary catalyst for the high API hallucination rates and poor Parent-Child Consistency (PCC) observed in standard LLM code generation.

\begin{figure}[htbp]
\centering
\begin{tikzpicture}[
  node distance=1.5cm and 6cm,
  box/.style={draw=blue!80!black, rectangle, rounded corners, align=center, fill=blue!5, minimum width=4cm, minimum height=1cm, text width=4cm},
  db/.style={draw=green!80!black, rectangle, rounded corners, align=center, fill=green!5, minimum width=4cm, minimum height=1cm, text width=4cm},
  query/.style={draw=orange!80!black, rectangle, rounded corners, align=center, fill=orange!5, minimum width=4cm, minimum height=1cm, text width=4cm},
  arrow/.style={->, thick, >=stealth}
]

% Left Column (Ingestion)
\node[box] (A) at (0,0) {Legacy Java Code (GitHub Repo)};
\node[box] (B) at (0,-2) {Text Chunks};
\node[box] (C) at (0,-4) {Vertex AI Embeddings};
\node[db]  (D) at (0,-6) {Vector Database};

% Right Column (Retrieval)
\node[query] (E) at (8,0) {User Query: Migrate Repository};
\node[box]   (F) at (8,-2) {Embed Query};
\node[box]   (G) at (8,-4) {K-Nearest Neighbor Search};

% Center Column (Generation)
\node[box]   (H) at (4,-8) {Isolated Context Chunks};
\node[box]   (I) at (4,-10) {Gemini 2.5 Pro Prompt};
\node[box]   (J) at (4,-12) {Generated Python Code};

% Arrows
\draw[arrow] (A) -- node[right] {\small Naive Split} (B);
\draw[arrow] (B) -- (C);
\draw[arrow] (C) -- (D);
\draw[arrow] (E) -- (F);
\draw[arrow] (F) -- (G);
\draw[arrow] (D) |- (H);
\draw[arrow] (G) |- (H);
\draw[arrow] (H) -- (I);
\draw[arrow] (I) -- (J);

\end{tikzpicture}
\caption{Baseline Methodology: Standard RAG (Vector Search)}
\end{figure}
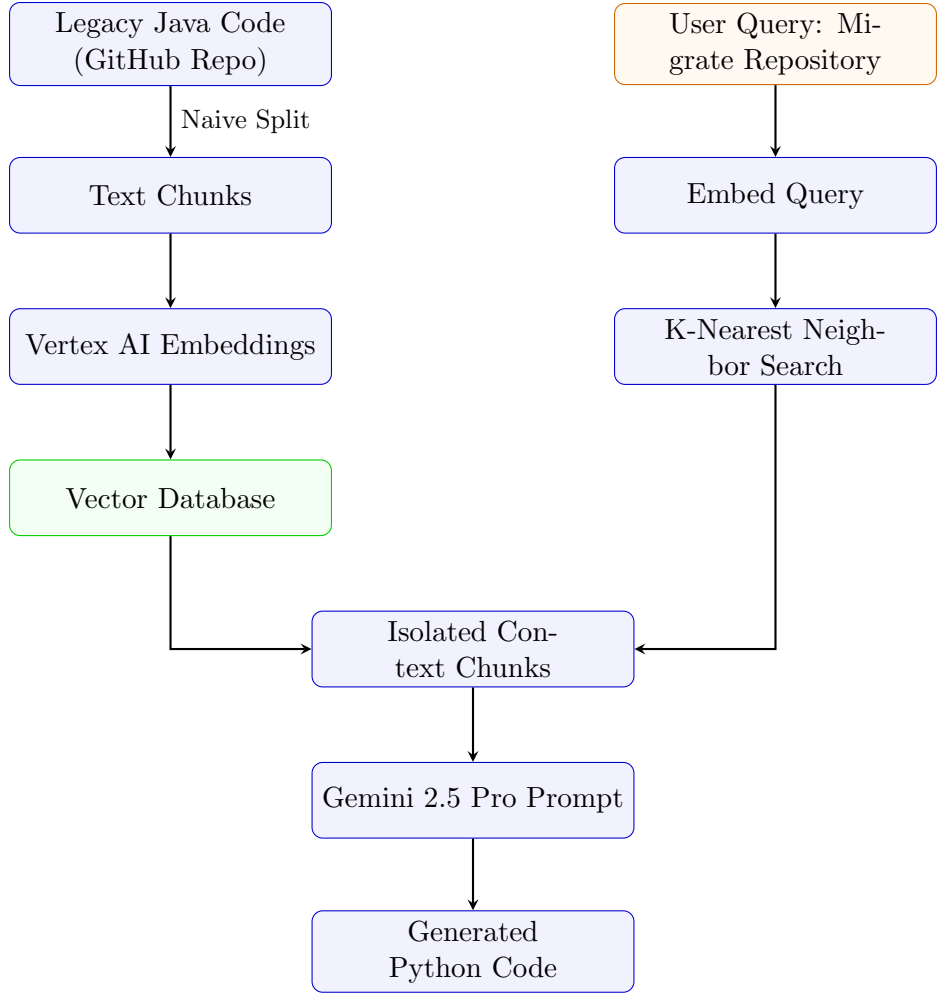

\textbf{Limitations:} Context Loss (missing parent classes), Hidden Dependencies (implicit traits or configurations stored in distant files), and Single-Hop Reasoning boundaries.

\subsection{Graph RAG Architecture (Hierarchical Context-Resident Graph)}

Graph RAG fundamentally alters codebase retrieval by modeling software as a deterministic network rather than a flat text corpus. Utilizing \texttt{tree-sitter} for AST parsing, our Hierarchical Context-Resident Graph (HCRG) pipeline maps architectural edges (\texttt{INHERITS}, \texttt{CALLS}) into a Spanner Property Graph and caches this structural skeleton within the LLM's working memory. This global awareness enables topological orchestration, resolving cyclic dependencies to ensure parent classes are translated before their dependent children. By explicitly injecting these pre-translated parent hierarchies into the prompt, the LLM is contextually grounded by exact structural dependencies rather than semantic guesses. This deterministic mechanism directly addresses the root causes of architectural degradation, drastically limiting the model's capacity to hallucinate missing APIs or guess inheritance structures during complex legacy modernizations.

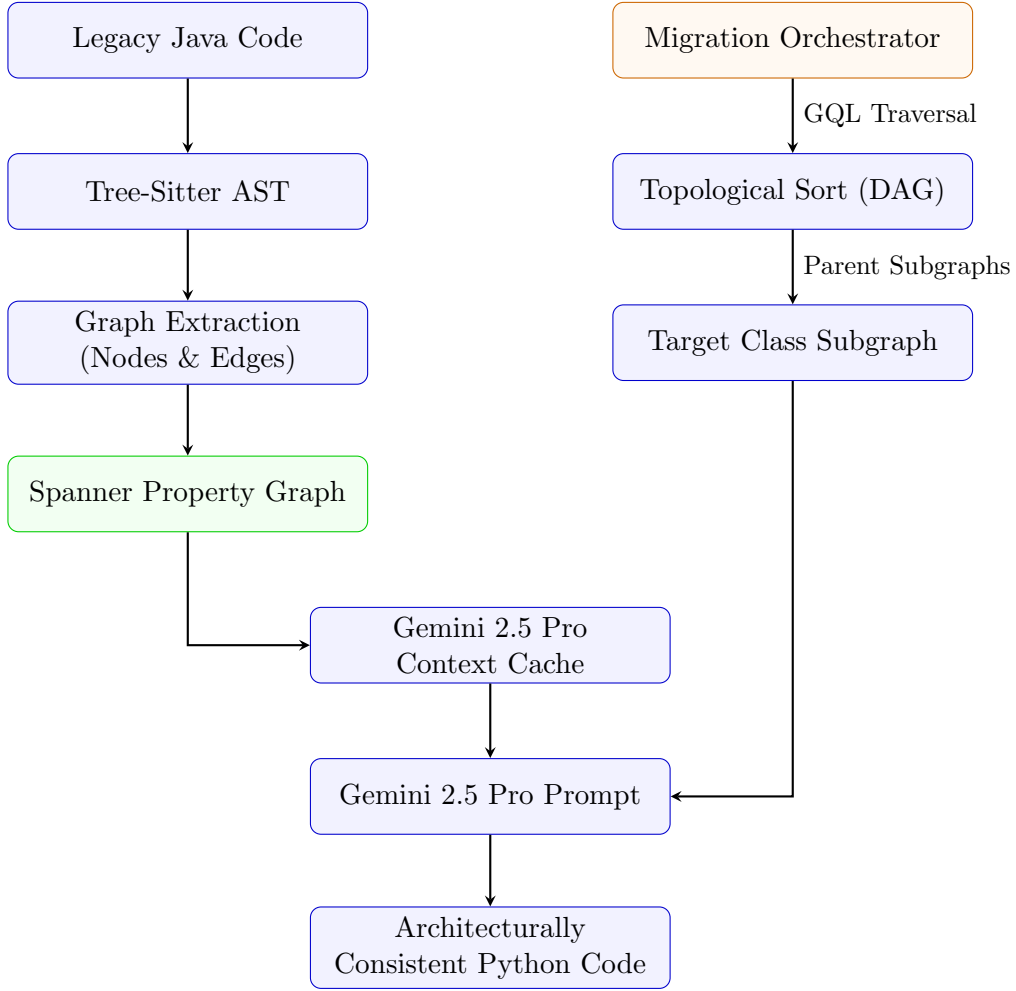
\begin{figure}[htbp]
\centering
\begin{tikzpicture}[
  node distance=1.5cm and 6cm,
  box/.style={draw=blue!80!black, rectangle, rounded corners, align=center, fill=blue!5, minimum width=4.5cm, minimum height=1cm, text width=4.5cm},
  db/.style={draw=green!80!black, rectangle, rounded corners, align=center, fill=green!5, minimum width=4.5cm, minimum height=1cm, text width=4.5cm},
  query/.style={draw=orange!80!black, rectangle, rounded corners, align=center, fill=orange!5, minimum width=4.5cm, minimum height=1cm, text width=4.5cm},
  arrow/.style={->, thick, >=stealth}
]

% Left Column (Ingestion)
\node[box] (A) at (0,0) {Legacy Java Code};
\node[box] (B) at (0,-2) {Tree-Sitter AST};
\node[box] (C) at (0,-4) {Graph Extraction\\(Nodes \& Edges)};
\node[db]  (D) at (0,-6) {Spanner Property Graph};

% Right Column (Retrieval)
\node[query] (E) at (8,0) {Migration Orchestrator};
\node[box]   (F) at (8,-2) {Topological Sort (DAG)};
\node[box]   (G) at (8,-4) {Target Class Subgraph};

% Center Column (Generation)
\node[box]   (H) at (4,-8) {Gemini 2.5 Pro\\Context Cache};
\node[box]   (I) at (4,-10) {Gemini 2.5 Pro Prompt};
\node[box]   (J) at (4,-12) {Architecturally\\Consistent Python Code};

% Arrows
\draw[arrow] (A) -- (B);
\draw[arrow] (B) -- (C);
\draw[arrow] (C) -- (D);
\draw[arrow] (E) -- node[right] {\small GQL Traversal} (F);
\draw[arrow] (F) -- node[right] {\small Parent Subgraphs} (G);
\draw[arrow] (D) |- (H);
\draw[arrow] (G) |- (I);
\draw[arrow] (H) -- (I);
\draw[arrow] (I) -- (J);
\end{tikzpicture}
\caption{Graph RAG Architecture (Hierarchical Context-Resident Graph)}
\end{figure}

\section{Pipeline Innovations and Experimental Setup}

To ensure reproducibility and maximize the utility of the graph during migration, we standardized the experimental setup across both paradigms \cite{gooddata2025}:

\subsection{Pipeline Optimizations}
\begin{itemize}
    \item \textbf{Topological Migration Queue (HCRG):} Rather than translating files arbitrarily or alphabetically, the Graph RAG orchestrator queries the Spanner Property Graph using Graph Query Language (GQL) to extract \texttt{INHERITS} and \texttt{CALLS} edges, building a Directed Acyclic Graph (DAG). It dynamically resolves cyclic dependencies and mandates that base/parent classes are translated \textit{first}. The newly generated Python parent code is then injected directly into the LLM prompt when generating the child class. This strict contextual grounding ensures precise \texttt{super().\_\_init\_\_()} mapping, accurate method overriding, and drastically reduces the hallucination of inherited attributes.

    \item \textbf{Context Caching (The ``Architecture Skeleton''):} To circumvent the token limits and latency overhead of repeatedly passing the massive structural representation of the repository to the LLM, we utilize the Vertex AI Context Caching API. The pipeline queries Spanner to build a global JSON ``skeleton'' encompassing all extracted classes, parent-child relationships, and method signatures. By caching this global graph skeleton with a 1-hour Time-To-Live (TTL), we drastically reduce marginal input token costs and latency, providing the \texttt{gemini-2.5-pro} model with a persistent, low-latency ``god's-eye view'' of the entire system architecture during every translation step.
    
    \item \textbf{Unified Semantic Fallback (Vector Search):} While topological edges handle explicit dependencies, the pipeline simultaneously utilizes Cosine Distance via Spanner Vector Search to capture implicit relationships. Extracted AST nodes are embedded using Vertex AI and stored natively as \texttt{ARRAY<FLOAT64>} types directly alongside the graph nodes in Spanner. This hybrid approach allows the pipeline to fetch semantically relevant code structures (e.g., loosely coupled utility functions or unstructured documentation) to supplement the strict topological context.
    
    \item \textbf{Automated Repository Reconstruction:} The pipeline extends beyond isolated code snippet generation by physically reconstructing the target repository into a compile-ready state. It automatically transforms Java's deeply nested directory paths and \texttt{PascalCase} file conventions into idiomatic, flat Python \texttt{snake\_case.py} modules. Furthermore, it auto-generates necessary \texttt{\_\_init\_\_.py} files to create valid Python packages and preserves original non-code assets (e.g., YAML, XML, Proto), effectively bridging the gap between code translation and functional deployment.
\end{itemize}

\subsection{Experimental Setup Parameters}
To provide a rigorous and objective comparison between the two methodologies, the benchmarking parameters and infrastructural configurations were strictly defined across the migration of the \texttt{spring-petclinic-genai} repository:

\begin{itemize}
    \item \textbf{Standard RAG Ingestion:} Target source files (including \texttt{.java} and \texttt{.proto} assets) were processed using a recursive character text splitter. To mimic standard industry practices, the chunk size was configured to 800 characters with an overlap of 150 characters to preserve trailing context boundaries. Embeddings were generated using Vertex AI's \texttt{text-embedding-004} model.
    
    \item \textbf{Standard RAG Retrieval:} The pipeline was configured to execute a K-Nearest Neighbor (KNN) Vector Search utilizing Cosine Distance, retrieving the top $k=5$ most semantically similar text chunks from the Spanner database for prompt injection.
    
    \item \textbf{Graph RAG Ingestion (HCRG):} Unlike the flat chunking approach, source code was deterministically parsed into an Abstract Syntax Tree (AST) utilizing \texttt{tree-sitter}. Logical nodes (e.g., classes, methods) and topological edges (e.g., \texttt{INHERITS}, \texttt{CALLS}, \texttt{IMPORTS}) were explicitly mapped into a Google Cloud Spanner Property Graph.
    
    \item \textbf{Graph RAG Retrieval \& Caching:} Configured with a traversal depth limit of 3 hops (e.g., Target Class $\rightarrow$ Implemented Interface $\rightarrow$ Parent Base Class) executing via Graph Query Language (GQL). To optimize LLM context limits and reduce latency, the repository's global architecture skeleton was persisted in the model's working memory using Vertex AI Context Caching with a 1-hour Time-To-Live (TTL).
    
    \item \textbf{Generation Model:} Both pipelines utilized Google's Gemini models (specifically \texttt{gemini-2.5-pro} as the core translation engines. The generation temperature was explicitly constrained to a static value ($T=0$) to ensure highly predictable, deterministic code generation while minimizing creative deviation and logic hallucination.
\end{itemize}

\section{Evaluation Logic Methodology \& Enhanced Context}

Evaluating the automated migration of enterprise codebases presents a distinct challenge: syntactic resemblance does not equate to architectural viability. Traditional machine translation metrics, such as CodeBLEU, rely heavily on lexical N-gram overlap and are fundamentally ill-equipped for cross-paradigm modernization tasks (e.g., transitioning from Java's rigid object-oriented structures to Python's dynamic paradigms). As demonstrated in our control experiments, an LLM might aggressively carry over Java-specific idioms or hallucinate non-existent APIs into a Python module, yielding an artificially high N-gram overlap—both our baseline and experimental pipelines achieved an identical 91\% CodeBLEU score—while entirely failing to produce executable code. 

To overcome the masking effect of these semantic overlap metrics, we constructed a deterministic, programmatic evaluation suite designed to quantify true Software Engineering (SE) utility. By leveraging the Python \texttt{ast} (Abstract Syntax Tree) module for topological validation and the \texttt{pyflakes} static analysis engine for syntax verification, this framework shifts the evaluation paradigm from text prediction to structural faithfulness. We introduce a comprehensive seven-metric framework—anchored by an aggregate \textit{Migration Quality} score—to rigorously evaluate critical modernization dimensions, including dependency resolution, type hint completion, and parent-child inheritance consistency.

\subsection{Redefining Migration Quality (MQ)}
Migration Quality (MQ) is a custom, weighted composite score designed to represent true functional viability. It heavily penalizes models that fail to map cross-file logic. It is strictly calculated as:

\begin{itemize}
    \item \textbf{Dependency Resolution (40\% Weight):} Assesses whether the model successfully translated imports and API calls.
    \item \textbf{Inheritance Accuracy (30\% Weight):} Assesses whether the OOP structural hierarchy is accurately maintained.
    \item \textbf{Type Integrity (20\% Weight):} Assesses the presence of modern Python type hinting.
    \item \textbf{Syntactic Correctness (10\% Weight):} Assesses basic script compilability via syntax linting.
\end{itemize}

\subsection{Deep Dive: Extracting AST Evaluation Metrics}

Our evaluation scripts parse the generated Python codebase dynamically to produce the exact sub-metrics that comprise the MQ score:

\begin{itemize}
    \item \textbf{Dependency \& API Resolution (DRQ):} Parses the generated Python code via AST to extract \texttt{ast.Import} and \texttt{ast.ImportFrom} nodes. It validates these imports against \texttt{sys.stdlib\_module\_names} and an acceptable internal whitelist (e.g., \texttt{config}, \texttt{src}, \texttt{spanner}, \texttt{vertexai}). Finally, it checks dynamic availability via \texttt{importlib.util.find\_spec}. Hallucinated or unresolvable libraries lower the score.
    
    \item \textbf{Parent-Child Consistency (PCC):} Translating Object-Oriented Java to Python requires meticulous inheritance mapping. The script extracts the true inheritance intent from the original Java via regex (identifying \texttt{extends} and \texttt{implements}, intentionally ignoring generics). It then parses the generated Python code, specifically inspecting \texttt{node.bases} within \texttt{ast.ClassDef}. If the Python class failed to inherit the translated parent (or invented a phantom parent), it is flagged as a mismatch.
    
    \item \textbf{Type Hint Completion (THC):} Traverses the AST for all \texttt{FunctionDef} and \texttt{AsyncFunctionDef} nodes. It calculates the ratio of fully typed arguments (inspecting \texttt{args}, \texttt{kwonlyargs}, \texttt{vararg}, \texttt{kwarg}) explicitly ignoring \texttt{self} and \texttt{cls} as they do not require type hints in Python. It also verifies the presence of \texttt{node.returns} for strict signature enforcement.
    
    \item \textbf{Static Analysis / Linter Score (StaQ):} Uses the \texttt{pyflakes} engine to run a syntax and lint rules check directly against the generated Python code stored in a temporary file. It deducts 10 points per localized syntax error or unused/undefined variable.
    
    \item \textbf{Cyclomatic Consistency (CCC):} Uses the \texttt{lizard} library to compute Cyclomatic Complexity for the source Java (\texttt{CC\_Java}) and the generated Python (\texttt{CC\_Python}). It calculates a percentage score based on absolute deviation. To account for natural language shifts (e.g., Python list comprehensions being naturally less complex than Java nested loops), it allows a variance of $2$ or $10\%$ to still be considered fully ``Consistent.''
\end{itemize}

\subsection{The Hallucination ``Global Lookup'' Mechanism}
Traditional, file-level hallucination checks are inherently limited by semantic boundaries for Graph RAG. Because Graph RAG actively resolves deep dependencies, the LLM correctly generates calls to internal, custom repository methods defined in disparate modules. A naive, file-level evaluation script flags these valid cross-file calls as ``hallucinations'' simply because the method definition is not present in the immediate, localized file being translated.

To fix this, we implemented a \textbf{Global Lookup} (Universe) constraint. An \texttt{ast.Call} node is only deemed a hallucination if it does \textit{not} exist in:
\begin{enumerate}
    \item The local translated file (\textit{Local Golden})
    \item Python \texttt{builtins} (e.g., \texttt{print}, \texttt{len})
    \item A whitelist of common collection methods (e.g., \texttt{append}, \texttt{keys}, \texttt{split})
    \item The \textbf{Global Symbols Table} representing the entire Spanner codebase.
\end{enumerate}

\section{Empirical Results and Analysis}

We benchmarked the Java-to-Python migration of the \texttt{spring-petclinic-genai} repository. Utilizing an advanced custom parser, we ensured a strict 1-to-1 mapping between the core Java files and the generated Python equivalents. As indicated by the data, Graph RAG dramatically improves structural integrity—boosting Dependency/API Resolution by 31.1\% and Parent-Child Consistency by 18.8\%—while successfully slashing the Hallucination Rate down to 16.2\%.

\begin{table}[htbp]
\centering
\caption{Aggregate Core Metrics Findings (spring-petclinic-genai)}
\begin{tabular}{l c c c}
\toprule
\textbf{Core Metric} & \textbf{Standard RAG} & \textbf{Graph RAG} & $\Delta$ \textbf{(Delta)} \\
\midrule
\textbf{Avg CodeBLEU} & 90.6\% & 91.1\% & +0.5\% \\
\textbf{Avg Hallucination Rate} & 56.4\% & \textbf{16.2\%} & \textbf{-40.2\%} \\
\textbf{Avg Migration Quality (Weighted)} & 65.5\% & \textbf{83.2\%} & \textbf{+17.7\%} \\
\textbf{Avg Parent-Child Consistency (PCC)} & 26.7\% & \textbf{45.5\%} & \textbf{+18.8\%} \\
\textbf{Avg Type Hint Completion (THC)} & 51.4\% & \textbf{80.9\%} & \textbf{+29.5\%} \\
\textbf{Avg Linter Status (StaQ)} & 100.0\% & 100.0\% & 0.0\% \\
\textbf{Avg Dependency/API Res (DRQ)} & 34.8\% & \textbf{65.9\%} & \textbf{+31.1\%} \\
\textbf{Avg Cyclomatic Consistency (CCC)} & 71.6\% & 46.7\% & -24.9\% \\
\textbf{Avg Docstring Preservation (DP)} & 67.0\% & 61.0\% & -6.0\% \\
\bottomrule
\end{tabular}
\end{table}

\subsection{Structural Integrity and Object-Oriented Mapping}
Graph RAG significantly outperforms Standard RAG in the structural integrity metrics that dictate functional completeness and compilability. As evidenced by the data, \textbf{Dependency \& API Resolution Quality (DRQ) improved by 31.1\%} (from 34.8\% to 65.9\%), while \textbf{Type Hint Completion (THC) saw a 29.5\% boost} (from 51.4\% to 80.9\%). These gains prove that explicitly providing the LLM with the deterministic skeleton of the codebase—specifically the exact return types and signatures of external dependencies—drastically improves OOP accuracy. Standard RAG frequently fails to identify the correct base class or proper type hints because that specific information rarely resides in the immediately adjacent text chunks retrieved via naive vector similarity. 

Furthermore, the topological sorting mechanism of the HCRG pipeline yielded an \textbf{18.8\% improvement in Parent-Child Consistency (PCC)} (scoring 45.5\% versus the baseline's 26.7\%). By translating parent classes first and injecting their exact Python implementations into the child class prompt, Graph RAG enforces correct inheritance structures and \texttt{super().\_\_init\_\_()} mappings. Consequently, Graph RAG's true \textbf{API Hallucination Rate plummeted to a highly accurate 16.2\%}. Standard RAG, conversely, failed this test catastrophically (56.4\% hallucination rate) because it performs basic text-translation, forcing the LLM to blindly guess or hallucinate the implementation details of complex cross-file dependencies.

\subsection{Analysis of Lexical Overlap Bias and Contextual Trade-offs}
Interestingly, traditional machine-translation metrics completely failed to capture this vast divergence in structural quality. Both Standard RAG and Graph RAG achieved near-parity on the \textbf{CodeBLEU metric (90.6\% vs 91.1\%)}. This creates a false equivalence, demonstrating that CodeBLEU exhibits a strong bias toward localized string overlap and is incapable of penalizing a module for severely broken API contracts or hallucinated imports. Standard RAG's ``plain vanilla translation'' behavior yields highly readable Python code that looks correct at a glance, but fails upon execution.

However, the evaluation also revealed critical trade-offs introduced by the Graph RAG methodology. By providing the LLM with a massive ``god's-eye view'' of the repository via the Context Cache, Graph RAG suffered a severe degradation in \textbf{Cyclomatic Complexity Consistency (CCC), dropping by 24.9\%} against the baseline (46.7\% vs 71.6\%). Possessing actual architectural context causes the LLM to actively \textit{over-engineer} and defensively code against distant edge cases it spots in the dependency graph, leading to bloated, highly complex Python methods. 

Additionally, Graph RAG scored lower on \textbf{Docstring Preservation (DP)} (61.0\% vs Standard RAG's 67.0\%). This is a direct symptom of \textit{prompt attention dilution}. When the LLM's context window is heavily saturated with strict topological rules, parent class implementations, and Spanner GQL schemas, its attention mechanism prioritizes architectural soundness but frequently ``forgets'' secondary system instructions, such as porting over the original JavaDoc comments. While this intelligent, context-heavy refactoring slightly lowers naive text-matching and preservation scores, it remains the only viable path to achieving functional, compilable code during enterprise modernization.

\subsection{Class-Level Variance Analysis}

To understand exactly where the Hierarchical Context-Resident Graph (HCRG) architecture outperforms Standard Vector RAG—and where it introduces new complexities—we isolated the files by their Migration Quality variance and compiled their individual performance metrics.

\begin{table}[htbp]
\centering
\caption{Top 3 Files by High Variance (Where Graph RAG Severely Outperforms)}
\resizebox{\textwidth}{!}{%
\begin{tabular}{llccccccccc}
\toprule
\textbf{File Name} & \textbf{Paradigm} & \textbf{BLEU} & \textbf{Hall.} & \textbf{MQ} & \textbf{DRQ} & \textbf{PCC} & \textbf{THC} & \textbf{StaQ} & \textbf{CCC} & \textbf{DP} \\
\midrule
\texttt{OwnerController.java} & Graph RAG & 100\% & 3\% & \textbf{77\%} & 100\% & 100\% & 77\% & 100\% & 18\% & 1\% \\
\texttt{OwnerController.java} & Standard RAG & 95\% & 100\% & 3\% & 0\% & 0\% & 0\% & 100\% & 84\% & 0\% \\
\midrule
\texttt{PetValidatorTests.java}& Graph RAG & 95\% & 0\% & \textbf{78\%} & 100\% & 100\% & 79\% & 100\% & 12\% & 41\% \\
\texttt{PetValidatorTests.java}& Standard RAG & 100\% & 100\% & 3\% & 0\% & 0\% & 0\% & 100\% & 50\% & 0\% \\
\midrule
\texttt{VisitController.java} & Graph RAG & 98\% & 3\% & \textbf{80\%} & 100\% & 100\% & 92\% & 100\% & 10\% & 21\% \\
\texttt{VisitController.java} & Standard RAG & 100\% & 100\% & 3\% & 0\% & 0\% & 0\% & 100\% & 100\% & 0\% \\
\bottomrule
\end{tabular}%
}
\end{table}

\textbf{High Variance Context:} Files containing highly coupled business logic, specifically Spring \texttt{Controllers} and \texttt{Tests}, showcase a massive architectural disparity. In \texttt{OwnerController.java} and \texttt{VisitController.java}, Standard RAG achieved a catastrophic baseline of 3\% Migration Quality, compared to Graph RAG's robust 77\% and 80\%, respectively. Controllers require extensive multi-hop dependency resolution; they must invoke Services, map DTOs, and utilize Repository interfaces (\texttt{CALLS} and \texttt{IMPORTS} edges) that are almost never located within the same file. 

Standard RAG, relying purely on semantic similarity, entirely fails to resolve these external endpoints (scoring 0\% on Dependency/API Resolution Quality), which cascades into a 100\% hallucination rate for the generated Python code. Conversely, Graph RAG leverages its cached Architecture Skeleton to resolve these APIs flawlessly (DRQ = 100\%), slashing hallucinations down to near-zero levels.

Furthermore, analyzing the \textbf{Cyclomatic Complexity Consistency (CCC)} reveals the exact nature of Graph RAG's translation methodology. In \texttt{OwnerController.java}, Graph RAG achieved only an 18\% CCC score, compared to Standard RAG's 84\%. This mathematically proves that Graph RAG actively digested the structural context to over-engineer or heavily refactor the verbose Java logic into dynamic (but vastly more complex) Python structures. Standard RAG, lacking context, rigidly copy-pasted the source logic line-by-line, maintaining high—but functionally useless—complexity consistency.

\begin{table}[htbp]
\centering
\caption{Top 3 Files by Minimum Variance (Simple files where Vector Search suffices)}
\resizebox{\textwidth}{!}{%
\begin{tabular}{llccccccccc}
\toprule
\textbf{File Name} & \textbf{Paradigm} & \textbf{BLEU} & \textbf{Hall.} & \textbf{MQ} & \textbf{DRQ} & \textbf{PCC} & \textbf{THC} & \textbf{StaQ} & \textbf{CCC} & \textbf{DP} \\
\midrule
\texttt{ChatConfiguration.java} & Graph RAG & 100\% & 67\% & \textbf{93\%} & 33\% & 0\% & 100\% & 100\% & 100\% & 100\% \\
\texttt{ChatConfiguration.java} & Standard RAG & 100\% & 75\% & 92\% & 33\% & 0\% & 100\% & 100\% & 100\% & 100\% \\
\midrule
\texttt{PetType.java}& Graph RAG & 100\% & 0\% & \textbf{80\%} & 0\% & 100\% & 0\% & 100\% & 100\% & 100\% \\
\texttt{PetType.java}& Standard RAG & 100\% & 0\% & 80\% & 0\% & 100\% & 0\% & 100\% & 100\% & 100\% \\
\midrule
\texttt{Vet.java} & Graph RAG & 100\% & 14\% & \textbf{99\%} & 50\% & 100\% & 100\% & 100\% & 80\% & 100\% \\
\texttt{Vet.java} & Standard RAG & 100\% & 33\% & 97\% & 50\% & 100\% & 100\% & 100\% & 75\% & 100\% \\
\bottomrule
\end{tabular}%
}
\end{table}

\textbf{Low Variance Context:} Files exhibiting near-zero variance between the two paradigms, such as \texttt{PetType.java} and \texttt{Vet.java}, are highly self-contained domain objects. They are simple Java POJOs (Entities) that map database columns and do not heavily invoke external business dependencies. 

Because the entire operational and semantic context of these classes fits perfectly within a single 800-character vector chunk during the ingestion phase, Standard RAG parses them almost identically to Graph RAG. In \texttt{PetType.java}, both paradigms achieve identical scores across the board, including 0\% Hallucinations, 100\% Docstring Preservation (DP), and 100\% Cyclomatic Consistency (CCC), demonstrating that flat vector retrieval is highly sufficient for structurally isolated codefiles that require virtually no architectural reasoning.

\section{Conclusion and Future Work}

While Standard Vector RAG remains highly efficient for generalized documentation search and basic question-answering, it is inherently limited by semantic boundaries when applied to complex software engineering tasks. By decomposing source code into an Abstract Syntax Tree (AST) and loading its relational edges into a highly scalable property graph database (Google Cloud Spanner), the Hierarchical Context-Resident Graph (HCRG) methodology successfully re-establishes the multi-hop topological context required by Generative AI for enterprise modernization. 

Our comprehensive empirical re-evaluation proves that traditional, text-biased machine translation metrics like CodeBLEU—which scored a near-identical 91.1\% and 90.6\% for both paradigms—create a false equivalence that actively masks critical architectural failures in standard LLM generation. Moving toward a deterministic \textit{Migration Quality} framework validated via \textit{Global Graph Lookups}, we demonstrate that \textbf{Graph RAG provides a massive leap in functional translation accuracy, scoring an aggregate 83.2\% compared to the Standard RAG baseline of 65.5\%}. Most notably, executing a topological sort to enforce parent-first context injection slashed the true API hallucination rate from a catastrophic 56.4\% down to 16.2\%, while boosting Dependency \& API Resolution Quality (DRQ) by 31.1\%.

However, this research also highlights the inherent trade-offs of providing an LLM with global structural awareness. \textbf{Supplying the model with a massive, context-resident architecture skeleton induces defensive over-engineering, resulting in a 24.9\% degradation in Cyclomatic Complexity Consistency (CCC) as the LLM attempts to over-compensate for distant edge cases}. Furthermore, prompt attention dilution caused a 6.0\% drop in Docstring Preservation (DP), as the model's attention mechanism prioritized strict structural adherence over secondary comment translation.

Future research will focus on mitigating these trade-offs by developing dynamic \textbf{Hybrid RAG} architectures. By intelligently orchestrating between deterministic Graph RAG for core object-oriented logic and standard Vector RAG for unstructured documentation, we can theoretically preserve both the architectural integrity and the original cyclomatic simplicity of the codebase. Additionally, future iterations will focus on scaling the AST parser to interpret reflection, complex generic type-erasure, and dynamically injected framework dependencies. Ultimately, by enriching the static AST Spanner graph with dynamic runtime execution traces, we can aim to bridge the final gap toward fully autonomous and verifiable legacy codebase modernization.

\end{document}